\documentclass[10pt,twocolumn,letterpaper]{article}

\usepackage[pagenumbers]{cvpr} 

\usepackage{graphicx}
\usepackage{amsmath}
\usepackage{amssymb}
\usepackage{booktabs}
\usepackage{mathtools}
\usepackage[x11names]{xcolor}
\usepackage{scalerel}
\usepackage{balance}
\usepackage{hhline}
\usepackage{multirow}
\usepackage{bm}
\usepackage{placeins} 
\usepackage{flafter}  
\usepackage{lipsum}
\usepackage{enumitem}
\setlist[itemize]{noitemsep, topsep=3pt}
\setlist[enumerate]{noitemsep, topsep=3pt}

\usepackage{setspace}

\usepackage{array}
\makeatletter
\newcommand{\thickhline}{%
    \noalign {\ifnum 0=`}\fi \hrule height 1pt
    \futurelet \reserved@a \@xhline
}
\newcolumntype{"}{@{\hskip\tabcolsep\vrule width 1pt\hskip\tabcolsep}}
\makeatother

\usepackage[ruled,linesnumbered]{algorithm2e}
\usepackage{etoolbox}

\makeatletter
\let\oldnl\nl
\newcommand{\nonl}{\renewcommand{\nl}{\let\nl\oldnl}}
\makeatother

\makeatletter
\patchcmd{\@algocf@start}
  {-1.5em}
  {0pt}
  {}{}
\makeatother

\usepackage[pagebackref,breaklinks,colorlinks]{hyperref}

\usepackage[capitalize]{cleveref}
\crefname{section}{Sec.}{Secs.}
\Crefname{section}{Section}{Sections}
\Crefname{table}{Table}{Tables}
\crefname{table}{Tab.}{Tabs.}

\def\cvprPaperID{2580} 
\def\confName{CVPR}
\def\confYear{2022}

\begin{document}

\title{HARA: A Hierarchical Approach for Robust Rotation Averaging}

\author{Seong Hun Lee\thanks{This work was partially supported by the Spanish govt. (PGC2018- 096367-B-I00) and the Arag{\'{o}}n regional govt. (DGA{\_}FSE T45{\_}20R).} \hspace{25pt} Javier Civera \\
I3A, University of Zaragoza, Spain\\
{\tt\small \{seonghunlee, jcivera\}@unizar.es}
\vspace{-1em}
}
\maketitle

\begin{abstract}
\vspace{-0.5em}
   We propose a novel hierarchical approach for multiple rotation averaging, dubbed HARA. 
   Our method incrementally initializes the rotation graph based on a hierarchy of triplet support.
   The key idea is to build a spanning tree by prioritizing the edges with many strong triplet supports and gradually adding those with weaker and fewer supports.
   This reduces the risk of adding outliers in the spanning tree. 
   As a result, we obtain a robust initial solution that enables us to filter outliers prior to nonlinear optimization.
   With minimal modification, our approach can also integrate the knowledge of the number of valid 2D-2D correspondences.
   We perform extensive evaluations on both synthetic and real datasets, demonstrating state-of-the-art results.
\end{abstract}

\vspace{-1em}
\section{Introduction}
We consider the problem of multiple rotation averaging in the presence of outliers, \ie, finding multiple absolute rotations $\mathbf{R}_i$ given a partial set of noisy, outlier-contaminated constraints on relative rotations $\mathbf{R}_{ij}=\mathbf{R}_i\mathbf{R}_j^\top$ \cite{hartley_2013_ijcv}.
This problem has direct application to structure-from-motion (SfM) \cite{martinec_2007_cvpr, enqvist_2011_iccv, arie_2012_ic3d, moulon_2013_iccv, wilson_2014_eccv, cui_2015_bmvc, cui_2015_iccv, ozyesil_2015_cvpr, cui_2017_cvpr, chen_2021_cvpr, zhu_2018_cvpr}, multiple point cloud registration \cite{govindu_2014_tip, tang_2015_cgf,arrigoni_2016_eccv, arrigoni_2018_cviu, huang_2019_cvpr,bhattacharya_2019_iccv, gojcic_2020_cvpr, moreira_2021_wacv} and simultaneous localization and mapping (SLAM) \cite{carlone_2015_icra, bourmaud_2014_accv, bourmaud_2016_eccv, bustos_2019_ral}.

In most global SfM pipelines, multiple rotation averaging is the \textit{de facto} standard for computing the initial orientations of the cameras:
After estimating the relative poses between image pairs (\eg, by matching feature descriptors such as SIFT \cite{sift} and running the 5-point algorithm \cite{nister_2004_PAMI} with RANSAC \cite{ransac}), one can solve the rotation averaging problem and obtain the absolute rotations with respect to a common reference frame.
These initial rotations are then used in subsequent operations such as translation estimation \cite{wilson_2014_eccv, cui_2015_bmvc}, pose graph optimization \cite{carlone_2015_icra, moreira_2021_wacv}, multiview triangulation \cite{lee_2019_bmvc, multiview_triangulation}
and bundle adjustment \cite{triggs_2000_bundle,  hartley_book, lee_2021_cvpr}.
As a result, all these tasks depend critically on the solution produced by the rotation averaging algorithm.

For this reason, numerous research endeavors have been made in the past decade to develop reliable and versatile rotation averaging methods.
However, even without any outliers in the input, solving a large-scale rotation averaging problem is nontrivial \cite{wilson_2016_eccv, wilson_2020_cvpr}. 
The problem only gets worse when the input contains outliers, which is often the case in practice \cite{wilson_2014_eccv, chatterjee_2018_tpami, moreira_2021_wacv}.
These outliers, if not handled properly, can easily degrade the estimation accuracy.

Commonly, rotation averaging is formulated as a nonlinear optimization problem and solved iteratively starting from some initial guess of the absolute rotations \cite{hartley_2011_cvpr, chatterjee_2018_tpami, shi_2020_icml, chen_2021_cvpr}.
If, however, this initial guess is severely affected by outliers, it becomes extremely difficult to obtain an accurate result later on.
Therefore, a robust initialization is essential for reliable rotation averaging in the presence of outliers.

In this work, we propose a novel method for robust multiple rotation averaging.
Our main contribution is a hierarchical initialization scheme that constructs a spanning tree of a rotation graph by propagating most reliable constraints first and less reliable ones later. 
We establish the hierarchy of reliability based on the number of consistent triplet constraints, as well as their level of consistency.
That is, we consider a constraint to be more reliable if it is strongly supported by many other constraints and less reliable if it has weaker or fewer supports.
Optionally, we can also incorporate the number of valid 2D-2D correspondences into the hierarchy.
Experimental results show that our approach can significantly improve the robustness of rotation averaging.
To download our code and the supplementary material, go to \url{https://seonghun-lee.github.io}.

\section{Related Work}
\label{sec:related}
Early works on motion averaging demonstrated various methods for estimating absolute rotations from pairwise constraints \cite{govindu_2001_cvpr, sharp_2002_eccv, fusiello_2002_eccv, govindu_2004_cvpr, martinec_2007_cvpr}.
In recent works, the focus has been on either (1) achieving the global optimality in the absence of outliers, or (2) obtaining a robust solution in the presence of outliers.
This work belongs to the second group.

\vspace{0.5em}
\noindent\textbf{(1) Globally optimal methods in outlier-free scenarios:}\\
In \cite{fredriksson_2012_accv, carlone_2015_iros}, globally optimal methods using Lagrangian duality are proposed.
In later works, more advanced optimization methods have been proposed to enhance the speed and scalability, while guaranteeing the global optimality of the solution \cite{eriksson_2018_cvpr, eriksson_2019_pami, rosen_2019_ijrr}.
For more recent works on optimal methods, we refer to \cite{dellaert_2020_shonan, parra_2021_cvpr, moreira_2021_iccv}.

\vspace{0.5em}
\noindent\textbf{(2) Robust methods in the presence of outliers:}\\
Various methods have been proposed to handle outliers (see \cite{tron_2016_cvpr} for a survey).
First, there are methods that attempt to detect and remove outliers, \eg, \cite{govindu_2006_accv, zach_2010_cvpr, crandall_2011_cvpr}.
Govindu \cite{govindu_2006_accv} uses a RANSAC-based method by sampling random spanning trees.
Zach \etal \cite{zach_2010_cvpr} employ a more tractable approach based on Bayesian inference from sampled loop inconsistencies. 
In \cite{crandall_2011_cvpr}, Crandall \etal use discrete belief propagation on a Markov random field to obtain the initial solution and remove edges with large errors.

On the other hand, some methods do not completely remove outliers, but instead suppress large errors during optimization. 
For example, Hartley \etal \cite{hartley_2011_cvpr} apply single rotation averaging under the $L_1$ norm to update each absolute rotation in a distributed manner.
Wang and Singer \cite{wang_2013_iijima} use semidefinite relaxation and an alternating direction method to minimize a cost function based on the $L_1$ norm.  
Chatterjee and Govindu \cite{chatterjee_2013_iccv, chatterjee_2018_tpami} apply the Lie-algebraic averaging \cite{govindu_2004_cvpr} using the iteratively reweighted least squares (IRLS) method with a robust loss function.
In \cite{arrigoni_2018_cviu}, Arrigoni \etal demonstrate that the spectral decomposition method in \cite{arie_2012_ic3d} can be robustified using the IRLS method.
Recently, Shi and Lerman \cite{shi_2020_icml} proposed an alternative optimization method, called message passing least squares, and demonstrated its advantages over the IRLS approach \cite{chatterjee_2018_tpami}.

Other robust methods directly model the presence of outliers in the optimization problem: 
Boumal \etal \cite{boumal_2013_cdc} take into account the outliers in the noise model and compute the maximum likelihood estimate via Riemannian trust-region optimization.
Arrigoni \etal \cite{arrigoni_2014_3dv, arrigoni_2018_cviu} intrinsically include the outliers in the cost function and estimate the rotations via low-rank and sparse matrix decomposition.

Another popular approach is to exploit additional visual information (\eg, the number of inlier feature matches or the similarity score) to identify inlier edges and obtain a robust initial solution \cite{enqvist_2011_iccv, tianwei_2016_eccv, cui_2018_icpr, chen_2021_cvpr, gao_2020_spl, gao_2021_ijcv}. 

Recently, learning-based approaches have been proposed in \cite{purkait_2020_eccv, yang_2021_cvpr}.
Although these supervised methods may not always generalize well to unfamiliar settings, they show impressive performance on data similar to the training data.

\section{Preliminaries and Notation}
We denote the Euclidean and the Frobenius norm of a 3D vector $\mathbf{v}$ by $\lVert\mathbf{v}\rVert$ and $\lVert\mathbf{v}\rVert_F$, respectively. 
We represent a rotation with a rotation matrix $\mathbf{R}\in SO(3)$ or a rotation vector $\mathbf{u}=\theta\widehat{\mathbf{u}}$
where $\theta$ and $\widehat{\mathbf{u}}$ are the angle and the unit axis of the rotation, respectively.
The two representations are related by Rodrigues' formula, and we denote the mapping between them by $\text{Exp}(\cdot)$ and $\text{Log}(\cdot)$ \cite{sola_2018_arxiv}:
\begin{equation}
    \label{eq:exp_log}
    \mathbf{R}=\mathrm{Exp}(\mathbf{u}), \quad
    \mathbf{u}=\mathrm{Log}(\mathbf{R}).
\end{equation}
In the context of SfM, the absolute rotation and translation of camera $i$ are denoted as $\mathbf{R}_i$ and $\mathbf{t}_i$, respectively.
Together, they transform a 3D point from the world frame to the camera reference frame: $\mathbf{x}_{i}=\mathbf{R}_i\mathbf{x}_w+\mathbf{t}_i$.
We denote with $\mathbf{R}_{jk}$ the relative rotation between $\mathbf{R}_j$ and $\mathbf{R}_k$, \ie, $\mathbf{R}_{jk}=\mathbf{R}_j\mathbf{R}_k^\top$.

\begin{figure}[t]
    \centering
    \includegraphics[width=0.41\textwidth]{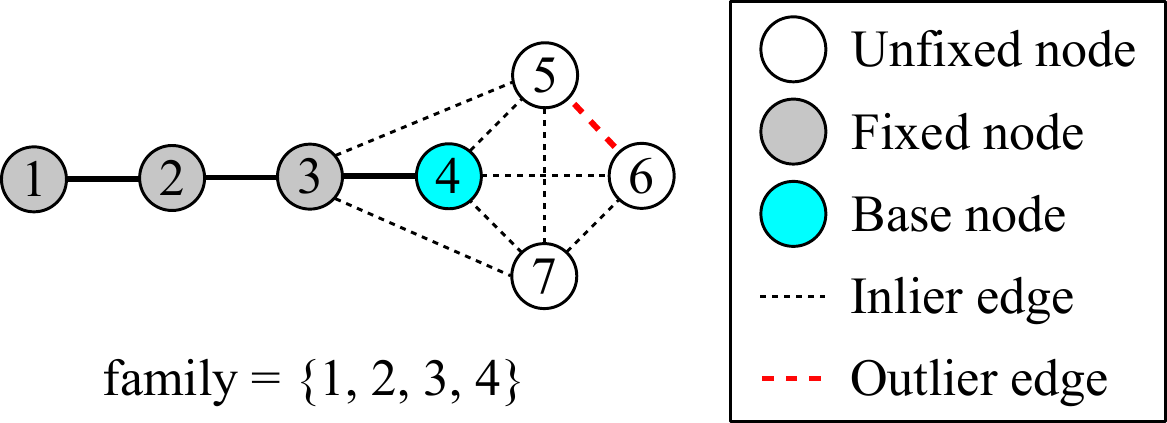}
    \vspace{-0.5em}
    \caption{
    In this example, the base node (4) has three non-family neighbors (5, 6 and 7).
    We check the triplet consistency (Eq. \eqref{eq:consistent_triplet}) without directly inferring the outlier edge:
    Node 5 has two triplet supports, \ie, (3, 4, 5) and (4, 5, 7), 6 has one support, \ie,  (4, 6, 7), and 7 has three supports, \ie, (3, 4, 7), (4, 5, 7) and (4, 6, 7).
    }
    \label{fig:triplet}
\end{figure}

The angular distance between $\mathbf{R}_j$ and $\mathbf{R}_k$ is defined as the angle of the rotation $\mathbf{R}_j\mathbf{R}_k^\top$, \ie,
\begin{equation}
    d(\mathbf{R}_j, \mathbf{R}_k) = \lVert\mathrm{Log}\left(\mathbf{R}_j\mathbf{R}_k^\top\right)\rVert.
\end{equation}
The chordal distance is related to the angular distance by the following equation \cite{hartley_2013_ijcv}:
\begin{align}
    d_\text{chord}(\mathbf{R}_j, \mathbf{R}_k) 
    :=& \lVert\mathbf{R}_j-\mathbf{R}_k\rVert_F \\
    =& 2\sqrt{2}\sin{\left(d(\mathbf{R}_j, \mathbf{R}_k)/2\right)}.
\end{align}
If both $\mathbf{R}_j$ and $\mathbf{R}_k$ have a small angle, their relative rotation can be approximated using the Baker-Campbell-Hausdorff (BCH) formula \cite{govindu_2004_cvpr}:
\begin{align}
    &\mathbf{R}_j\mathbf{R}_k^\top \approx \mathrm{Exp}(\mathbf{u}_j - \mathbf{u}_k).\\
    \Rightarrow \ & \mathrm{Log}\left(\mathbf{R}_j\mathbf{R}_k^\top\right) \approx \mathbf{u}_j - \mathbf{u}_k. \label{eq:bch}
\end{align}
In the following, we list some important terminology:
\begin{itemize}[leftmargin=*]
    \item \textbf{Nodes and edges:}
    Multiple absolute rotations are related to each other in pairs, so the underlying structure can be represented by a graph.
    In this context, the \textit{nodes} represent the unknown absolute rotations, and the \textit{edges} represent the known pairwise constraints.

    \item \textbf{Neighbors:}
    When two nodes are connected by an edge, they are each other's \textit{neighbors}.
    
    \item \textbf{Fixed nodes and family:} 
    Once a node is initialized with some absolute rotation, we call it \textit{fixed}. 
    A \textit{family} refers to the set of all fixed nodes.
    The goal of the initialization is to have all nodes included in the family.
    
    \item \textbf{Base node:}
    One of the fixed nodes can be chosen as the \textit{base node} at any time during the initialization.
    This is the node from which the yet-incomplete spanning tree will branch out if a certain condition is met.
    
    \item \textbf{Consistent triplet:}
    Node $i$, $j$ and $k$ form a \textit{consistent triplet} if and only if the input relative rotations satisfy
    \begin{equation}
        \label{eq:consistent_triplet}
        d_\text{chord}(\mathbf{R}_{ij}^\text{in}, \mathbf{R}_{ik}^\text{in}\mathbf{R}_{kj}^\text{in}) < \epsilon,
    \end{equation}
    where $\epsilon$ is called a \textit{loop threshold}.
    A triplet that satisfies Eq. \eqref{eq:consistent_triplet} under small $\epsilon$ is described as ``strong", and one that does it under relatively large $\epsilon$ is described as ``weak". 
    If a triplet contains one or more outlier edges, it is most likely to be inconsistent and fail to meet Eq. \eqref{eq:consistent_triplet}.
    
    \item \textbf{Number of triplet supports:}
    Suppose that a base node has several non-family neighbors, including node $i$.
    The number of \textit{triplet supports} of neighbor $i$ refers to the number of consistent triplets formed by the base node, node $i$ and another neighbor of the base node.
    A simple example is illustrated in Fig. \ref{fig:triplet}.
\end{itemize}

\newpage

\begin{figure}[t]
    \centering
    \includegraphics[width=0.38\textwidth]{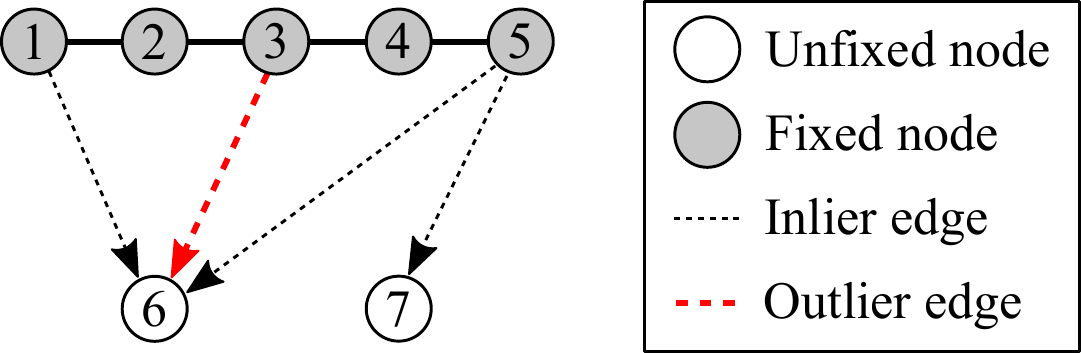}
    \vspace{-0.5em}
    \caption{
    Here, no matter which family member is chosen as the base node, we cannot form a consistent triplet. 
    In this case, we let every family member vote for their non-family neighbors and add the one with the most votes (node 6) to the family.
    Among the candidate rotations propagated from node 1, 3 and 5, we choose the one that is closest to their robust average (obtained using \cite{robust_single_rotation_averaging}).
    }
    \label{fig:voting}
\end{figure}

\begin{algorithm}[t]
\setstretch{1}
\caption{Hierarchical Initialization (simplified)}
\label{alg:simplified}
\small
 $s\gets s_{\scaleto{\text{init}}{5pt}}$, $\texttt{family}\gets\{\}$, $\texttt{newFamily}\gets\{\}$\; 
Add the node with the most neighbors to \texttt{family} and \texttt{newFamily}, and set its rotation to identity\;
\While{not all nodes are in \texttt{family}}{
\While{\texttt{newFamily} is not empty}
{
    Choose a member of \texttt{newFamily} as the base node and remove it from \texttt{newFamily}\label{line:choosing_base1}\;
    Propagate away from the base node to its non-family neighbors using Eq. \eqref{eq:propagation}, and add those with $s$ or more triplet supports to \texttt{family} and \texttt{newFamily}\;
    \If{at least one node is added to \texttt{newFamily}}
    {
    $s\gets s_{\scaleto{\text{init}}{5pt}}$\;
    }
}
For the next base node, choose the family member that has the most non-family neighbors with $s$ or more supports\label{line:counting}\;
\eIf{the base node has at least one non-family neighbor with $s$ or more supports}
{   
    Add the base node to \texttt{newFamily}.
}
{
    $s\gets s - 1$\;
}

\If{$s = 0$}
{
    Let every family member vote for their non-family neighbors, add the one with the most votes to \texttt{family} and \texttt{newFamily}, and set its rotation via single rotation averaging (see Fig. \ref{fig:voting})\;
    $s\gets s_{\scaleto{\text{init}}{5pt}}$\;
}

}
\end{algorithm}

\section{Method}
\label{sec:method}
The proposed method consists of three steps:
\begin{enumerate}
    \item Robust initialization of the absolute rotations by building a spanning tree in a hierarchical manner. 
    If the number of inlier matches is known for all edges, we can optionally incorporate it in the initialization.
    \item Filtering the edges that do not conform to the initial solution to remove as many outliers as possible.
    \item Iterative local refinement using nonlinear optimization.
\end{enumerate}
To make the initialization step easier to understand, we first describe the simplified version in Section \ref{subsec:initialization_simplified} and then the full version in Section \ref{subsec:initialization_full}.
We explain the edge filtering and the local refinement in Section \ref{subsec:edge_filtering} and \ref{subsec:refinement}, respectively.

\subsection{Hierarchical initialization (simplified version)}
\label{subsec:initialization_simplified}
We initialize the absolute rotations by constructing a spanning tree of the graph.
As we expand the tree incrementally, we want to avoid as many outlier edges as possible, so  we start adding the most reliable edges first.
In our method, there are two modes of tree expansion: (1) based on the triplet support, or (2) via single rotation averaging.

First, we set a certain integer threshold $s$ (called a \textit{support threshold}) and check if the base node has any non-family neighbors with $s$ or more triplet supports.
If so, we add these neighbors to the family and obtain their rotations $(\mathbf{R}_\text{N}^\text{est})$ by propagating from the base node $(\mathbf{R}_\text{B}^\text{est})$, \ie,
\begin{equation}
\label{eq:propagation}
    \mathbf{R}_\text{N}^\text{est} \gets \mathbf{R}_\text{NB}^\text{in}\mathbf{R}_\text{B}^\text{est}.
\end{equation}
For example, if $s$ = 2 in Fig. \ref{fig:triplet}, we would add node 5 and 7 in the family, but not 6.
If all non-family neighbors of the family have fewer than $s$ triplet supports, we update $s\gets s-1$ and, for the next base node, choose the family member that has the most non-family neighbors with $s$ or more supports.
We repeat the same propagation process afterwards.

Another way to expand the tree is to add a node via single rotation averaging when all non-family neighbors of the family have zero triplet support.
In this case, we let every family member vote for their non-family neighbors, and the one with the most votes is added to the family.
This node also becomes the next base.
To determine its rotation, we first obtain the candidate rotations by propagating from the family nodes that voted for it.
Then, we average these rotations using the robust single rotation averaging method in \cite{robust_single_rotation_averaging}.
Finally, the candidate rotation that is closest to the result is assigned to the node.
Fig. \ref{fig:voting} shows an example.

\begin{figure*}[t]
    \centering
    \includegraphics[width=0.98\textwidth]{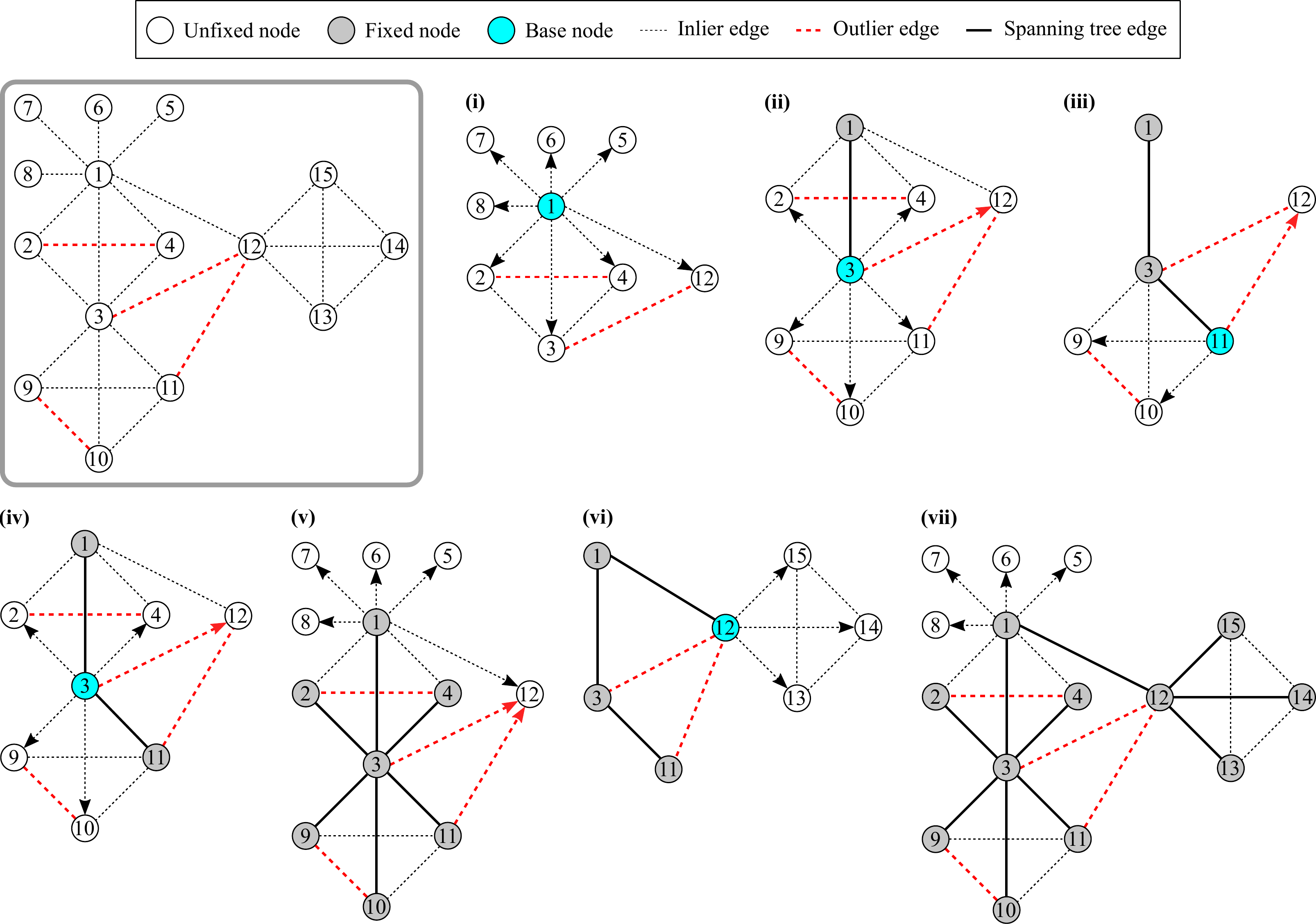}
    \caption{
    \textbf{[Top left]} A toy example.
    We show the steps of our initialization algorithm with $s = 2$ and a fixed loop threshold.\\
    \phantom{aaa}\textbf{(i)} First, we choose the node with the most neighbors (node 1) as the base node and set its rotation to identity.
    This node is the first member of the family.
    By propagating away from this node using Eq. \eqref{eq:propagation}, its neighbors get tentative rotations.
    For each neighbor, we count the number of triplet supports (\ie, the number of other neighbors supporting it), and if it has $s$ or more supports, we add it to the family.
    In this example, node 3 is supported by two other neighbors (node 2 and 4), and it is the only one added to the family.
    Node 3 becomes the next base node, and we fix its rotation.
    Also, the edge (1, 3) turns into a spanning tree edge.\\   
    \phantom{aaa}\textbf{(ii)} We repeat the same process by propagating away from the new base node (node 3).
    The only non-family neighbor that has $s$ or more supports is node 11, so we fix its rotation, add it to the family, and select it as the next base node.\\
    \phantom{aaa}\textbf{(iii)} We propagate away from the new base node (node 11), but no neighbor has enough supports.
    In this case, we update $s\gets s-1$ and check for each family member how many neighbors have $s$ or more supports: node 1 has two (node 2 and 4), node 3 has four (node 2, 4, 9, 10), and node 11 has two (node 9 and 10).
    Since node 3 has the most, it becomes the next base node.
    Note that in the full version of the algorithm, we do this counting as soon as the base node changes and store the results for reuse (more details in Section \ref{subsec:initialization_full}).\\
    \phantom{aaa}\textbf{(iv)} We propagate away from the new base node (node 3), and the non-family neighbors with $s$ (= 1) supports are node 2, 4, 9 and 10.
    These four nodes are added to the family, and their rotations are fixed.\\
    \phantom{aaa}\textbf{(v)} With node 1--4 and 9--11 in the family, none of their non-family neighbors has a single support.
    In this case, each family member votes for their non-family neighbors, and the one with the most votes is added to the family.
    This node (node 12) also becomes the next base node.
    To determine its rotation, we first average the candidate rotations propagated from node 1, 3 and 11 using \cite{robust_single_rotation_averaging}.
    Then, the candidate rotation that is closest to the result is assigned to node 12, and the corresponding edge becomes a spanning tree edge.
    In this example, let us suppose that it is the edge (1, 12).\\
    \phantom{aaa}\textbf{(vi)} Every time a node is added to the family, we reset $s$ to the initial value ($s\gets2$).
    Afterwards, we repeat the process of propagating away from the base node and adding the neighbors with $s$ or more supports to the family.
    In this example, node 12 has three non-family neighbors (node 13, 14, 15) and they all have $s$ supports.
    Therefore, all three of them are added to the family and their rotations are fixed.\\
    \phantom{aaa}\textbf{(vii)} With node 1--4 and 9--13 in the family, none of their non-family neighbors has $s$ (= 2) supports. 
    We update $s\gets s-1$ and check again, but none has a single support. 
    Now, as in Step (v), we let every family member vote for their non-family neighbors, and add the one with the most votes. 
    Repeating this procedure adds node 5--8 to the family one by one.
    Finally, all nodes are in the family and their rotations are fixed.
    The algorithm returns the estimated rotations of all nodes.
    }
    \label{fig:toy}
\end{figure*}

At each iteration, our initialization algorithm decides between the two aforementioned modes of tree expansion.  
We first try expanding based on the triplet support by adapting the support threshold $s$, and when $s = 0$, we expand the tree via voting and single rotation averaging.
Every time a node is added to the family, we reset $s$ to the initial value.
Alg. \ref{alg:simplified} summarizes the procedure and Fig. \ref{fig:toy} shows a toy example.

\clearpage

\clearpage
\newpage

\begin{algorithm}[t]
\setstretch{1}
\caption{Hierarchical Initialization (full version)}
\label{alg:full}
\small
 $s\gets s_{\scaleto{\text{init}}{5pt}}$, $\texttt{family}\gets\{\}$, $\texttt{newFamily}\gets\{\}$\; 
Add the node with the most neighbors to \texttt{family} and \texttt{newFamily}, and set its rotation to identity\;

\colorbox{DarkOliveGreen1}{\parbox{0.7\linewidth}{Determine the loop thresholds $\epsilon_1$, $\epsilon_2$, \dots, $\epsilon_m$\; }}\label{line:epsilon}

\colorbox{DarkOliveGreen1}{\parbox{0.24\linewidth}{$i\gets1, \ \epsilon\gets\epsilon_{\scaleto{i}{5pt}}$\; }}

\colorbox{DarkOliveGreen1}{\parbox{0.82\linewidth}{$\texttt{snTable}\gets\text{Zero 3D array of dimension } n\times m \times s$\;}}

\nonl ($n$ is \#nodes, $m$ is \#loop thresholds, $s$ is a  support threshold)

\While{not all nodes are in \texttt{family}\label{line:outer_start}}{
\While{\texttt{newFamily} is not empty}
{
    Choose a member of \texttt{newFamily} as the base node and remove it from \texttt{newFamily}\label{line:choosing_base2}\;
    Propagate away from the base node to its non-family neighbors  using Eq. \eqref{eq:propagation} and add those with $s$ or more triplet supports to \texttt{family} and \texttt{newFamily}\;
    
    \colorbox{DarkOliveGreen1}{\parbox{0.55\linewidth}{Update \texttt{snTable} for the base node\label{line:update_snTable}\;}}
    
    \If{at least one node is added to \texttt{newFamily}}
    {
    $s\gets s_{\scaleto{\text{init}}{5pt}}$, 
    \colorbox{DarkOliveGreen1}{\parbox{0.24\linewidth}{$i\gets1, \ \epsilon\gets\epsilon_{\scaleto{i}{5pt}}$\; }}
    }
}
\colorbox{DarkOliveGreen1}{\parbox{0.88\linewidth}{In \texttt{snTable}, find the family member that has the most non-family neighbors with $s$ or more triplet supports under the current threshold $\epsilon$. 
Choose it as the base node\label{line:use_snTable}\;}}

\eIf{the base node has at least one non-family neighbor with $s$ or more supports}
{   
    Add the base node to \texttt{newFamily}.
}
{
    
    \eIf{$i < m$\label{line:adapt_start}}
    {
    \colorbox{DarkOliveGreen1}{\parbox{0.3\linewidth}{$i\gets i+1, \ \epsilon\gets\epsilon_{\scaleto{i}{5pt}}$\; }}
    }
    {
    \colorbox{DarkOliveGreen1}{\parbox{0.43\linewidth}{$s\gets s-1, \ i\gets 1, \ \epsilon\gets\epsilon_{\scaleto{i}{5pt}}$\; }}
    }\label{line:adapt_end}
}

\If{$s = 0$}
{
    Let every family member vote for their non-family neighbors, add the one with the most votes to \texttt{family} and \texttt{newFamily}, and set its rotation via single rotation averaging (see Fig. \ref{fig:voting})\label{line:vote}\;
    $s\gets s_{\scaleto{\text{init}}{5pt}}$, \colorbox{DarkOliveGreen1}{\parbox{0.24\linewidth}{$i\gets1, \ \epsilon\gets\epsilon_{\scaleto{i}{5pt}}$\; }}
}

}\label{line:outer_end}
\end{algorithm}

\subsection{Hierarchical initialization (full version)}
\label{subsec:initialization_full}
The simplified algorithm described in the previous section constructs a spanning tree by adding the most supported edges first.
In the full version, we consider two more aspects: the loop threshold $\epsilon$ in Eq. \eqref{eq:consistent_triplet}, and optionally, the number of valid 2D-2D correspondences.
We highlight the differences between the two versions in Alg. \ref{alg:full}.

In the simplified version, the consistency of a triplet depends entirely on a single threshold $\epsilon$ we set.
In the full version, we set multiple thresholds ($\epsilon_1$, $\epsilon_2$, \dots, $\epsilon_m$ in ascending order) and adaptively switch between them. 
Specifically, we start from the smallest (the strictest) threshold and gradually move on to larger (less strict) thresholds. 
As a result, the following hierarchy is established:
\begin{enumerate}[leftmargin=*]
    \item Neighbor nodes with many triplet supports under small $\epsilon_i$ are added to the family first.
    \item Those with many supports under large $\epsilon_i$ are added next.
    \item Those with few supports under small $\epsilon_i$ are added next.
    \item Those with few supports under large $\epsilon_i$ are added last. 
\end{enumerate}

Another change from the simplified version is that we store the number of supported neighbors each time we try propagating away from the base node. 
This data is stored in a \textit{supported neighbors table} (SN table), a 3D array whose dimensions correspond to the base node index, the threshold index, and the number of triplet supports.
We organize this table such that the entry at position ($x$, $y$, $z$) corresponds to the number of non-family neighbors of base node $x$ that have $z$ or more supports under the $y$-th threshold $\epsilon_y$.
Note that each time we update the SN table for the current base node, we update it for all $y = 1, 2, \dots, m$ and $z = 1, 2, \dots, s_{\scaleto{\text{init}}{5pt}}$.
This is done in line \ref{line:update_snTable} of Alg. \ref{alg:full}.

The advantage of maintaining this table is that we can reuse it to promptly find the node with the most number of supported neighbors for any given $s$ and $\epsilon$ (line \ref{line:use_snTable} of Alg. \ref{alg:full}).
This operation is necessary when we have to choose the next base node after $s$ is decremented.
Although the data in the SN table may sometimes be outdated (because some non-family neighbors can turn into family members later), we can at least avoid having to evaluate the neighbors of all family members repeatedly (\ie, line \ref{line:counting} of Alg. \ref{alg:simplified}).

Our approach can also seamlessly integrate the knowledge of the number of valid 2D-2D correspondences. 
This can be done with minimal modification of Alg. \ref{alg:full}:
Let $d_1$, $d_2$, \dots, $d_k$ (in descending order) be some thresholds we set for the number of valid 2D-2D correspondences.
Then, we run the outer loop (line \ref{line:outer_start}--\ref{line:outer_end}) while pretending that all edges whose valid correspondences are fewer than $d_1$ do not exist.
When the number of total votes becomes zero in line \ref{line:vote}, we reset $s$, $\epsilon$ and the SN table to the initial state, switch the correspondence threshold to the next one ($d_2$) and continue.  
This process ensures that the edges with very few valid correspondences are added last.

\noindent\textbf{Implementation details:}
\begin{enumerate}[leftmargin=*]
    \item  In line \ref{line:choosing_base1} of Alg. \ref{alg:simplified} and line \ref{line:choosing_base2} of Alg. \ref{alg:full}, if \texttt{newFamily} has multiple members, we choose the one with the most neighbors as the base node.
    This is because we want to add well-connected nodes first to minimize drift.
    \item In all of our experiments in this paper, we fix $s_{\scaleto{\text{init}}{5pt}} = 10$, $d_1 = 5$ and $d_2 = 0$.
    \item For good performance, the loop thresholds should reflect the noise level of the inlier edges.
To this end, we use a simple heuristic method to determine their values in line \ref{line:epsilon} of Alg. \ref{alg:full}:
For each edge $(i, j)$, we sample at most 10 common neighbors of node $i$ and $j$, forming up to 10 triplets $(i, j, k)$.
We compute the loop errors \eqref{eq:consistent_triplet} of all triplets from all edges and collect only those below $1$.
Then, we set the loop thresholds $\epsilon_1$, $\epsilon_2$ and $\epsilon_3$ to the 10th, 20th and 30th percentile of the collected errors.
\end{enumerate}

\subsection{Edge filtering}
\label{subsec:edge_filtering}
Having obtained an initial solution $(\mathbf{R}_i^\text{est}$ for $i$ = 1, 2, $\dots, n)$ from the spanning tree in Section \ref{subsec:initialization_full}, we next filter potential outlier edges in the full rotation graph before optimizing the solution.
This is done by checking whether or not each edge conforms to the initial solution.
Specifically, we consider edge $(j,k)$ as an outlier and exclude it from the further operations if the following condition is met:
\begin{equation}
    d_\text{chord}\left(\mathbf{R}_{jk}^\text{in}, \mathbf{R}_j^\text{est}{\mathbf{R}_k^\text{est}}^\top\right) > \tau,
\end{equation}
where $\tau$ is some threshold (we set $\tau=1$ in this work).

While this filtering step often enhances the robustness for moderate outlier ratios ($< 0.3$), we found that it sometimes worsens the accuracy for higher outlier ratios.
Therefore, we skip this step when we deem the outlier ratio to be too high. 
In practice, we assume that this is the case when the median of the loop errors from all sampled triplets is larger than $1$ (see the implementation detail 3 of Section \ref{subsec:initialization_full}).

\subsection{Local refinement}
\label{subsec:refinement}
Given the initial solution and the filtered constraints $(\mathbf{R}_{jk}^\text{in} \text{ for } (j,k)\in \text{ filtered edges})$, we perform an iterative local refinement using the optimization method proposed in \cite{chatterjee_2018_tpami}.
In the following, we briefly summarize this method.

The goal here is to find the optimal updates such that the updated solution fits the constraints better, \ie,
\begin{equation}
    \mathbf{R}_{jk}^\text{in} = \left(\mathbf{R}_j^\text{est}\Delta\mathbf{R}_j\right)\left(\mathbf{R}_k^\text{est}\Delta\mathbf{R}_k\right)^\top.
\end{equation}
Rearranging this and taking the $\mathrm{Log}$ \eqref{eq:exp_log} of both sides gives
\begin{equation}
    \mathrm{Log}\left(\mathbf{R}_j^{\text{est}\top}\mathbf{R}_{jk}^\text{in}\mathbf{R}_k^\text{est}\right) = \mathrm{Log}\left(\Delta\mathbf{R}_j\Delta\mathbf{R}_k^\top\right).
\end{equation}
Assuming that the updates are small, we can use the approximation in Eq. \eqref{eq:bch} on the right-hand side and obtain
\begin{equation}
    \label{eq:constraint}
    \mathrm{Log}\left(\mathbf{R}_j^{\text{est}\top}\mathbf{R}_{jk}^\text{in}\mathbf{R}_k^\text{est}\right) \approx \Delta\mathbf{u}_{j} - \Delta\mathbf{u}_{k},
\end{equation}
where $\Delta\mathbf{u}_{j}$ and $\Delta\mathbf{u}_{k}$ are the rotation vectors of $\Delta\mathbf{R}_j$ and $\Delta\mathbf{R}_k$, respectively.
Since the left-hand side of Eq. \eqref{eq:constraint} is known, stacking these equations for all filtered edges results in a linear system of equations, which we solve using a linear algebra library.
We update the rotations, \ie, $\mathbf{R}_i^\text{est}\gets\mathbf{R}_i^\text{est}\Delta\mathbf{R}_i$ for all $i$, plug them back into Eq. \eqref{eq:constraint} and repeat the same process until convergence.
In practice, we carry out the optimization using the IRLS method with the $\ell_{\frac{1}{2}}$ loss function, as in \cite{chatterjee_2018_tpami}.
Also, to reduce the total number of arithmetic operations, all rotations (both absolute and relative) are parameterized as quaternions.
For more details, we refer to the original work \cite{chatterjee_2018_tpami}.

\newpage
\section{Results}
\label{sec:results}
We compare our method with the following methods:
R-GoDec\footnote{\url{http://www.diegm.uniud.it/fusiello/demo/gmf/}} \cite{arrigoni_2014_3dv, arrigoni_2018_cviu}, Eig-IRLS\footnote{The code was kindly provided by the authors of \cite{arrigoni_2018_cviu}.} \cite{arrigoni_2018_cviu}, IRLS-$\ell_{\frac{1}{2}}$\footnote{\url{http://www.ee.iisc.ac.in/labs/cvl/research/rotaveraging/}\label{url:chatterjee}} \cite{chatterjee_2018_tpami}, MPLS\footnote{\url{https://github.com/yunpeng-shi/MPLS}} \cite{shi_2020_icml} and Hybrid RA\footnote{\url{https://github.com/AIBluefisher/GraphOptim}} \cite{chen_2021_cvpr}.
Since the implementation of the view graph filtering (VGF) in Hybrid RA is not publicly available, we reproduced this part by ourselves.
Note that this part is only applicable if 2D-2D correspondences are given for all edges.
All methods are implemented in MATLAB, except Hybrid RA which is written in C++.
We run all methods on a laptop with Intel's 4th Gen i7 CPU (2.8 GHz).

We evaluate the accuracy using two error metrics:
\vspace{-0.2em}
\begin{align}
    \theta_{1} &= \min_{\mathbf{R}_{\scaleto{\textrm{{align}}}{6pt}}}\frac{1}{n}\sum_{i=1}^n d\left(\mathbf{R}_i^\text{gt}, \mathbf{R}_i^\text{est}\mathbf{R}_\text{align}\right),\label{eq:theta_1}\\
    \theta_{2} &=
    \min_{\mathbf{R}_{\scaleto{\textrm{{align}}}{6pt}}}\sqrt{\frac{1}{n}\sum_{i=1}^n d\left(\mathbf{R}_i^\text{gt},\mathbf{R}_i^\text{est}\mathbf{R}_\text{align}\right)^2}.\label{eq:theta_2}
\end{align}
They respectively represent the optimal mean and RMS error after aligning the estimated rotations to the ground truth.
The rotation $\mathbf{R}_\text{align}$ in Eq. \eqref{eq:theta_1} and \eqref{eq:theta_2} can be obtained by solving the single rotation averaging problem under the $L_1$ and $L_2$ norm, respectively \cite{hartley_2011_cvpr, hartley_2013_ijcv}.

\subsection{Synthetic data}
\label{subsec:synthetic}
For a controlled study of various factors, we run Monte Carlo simulations in multiple settings:
We generate $n$ random rotations in a circular order and obtain the relative rotation of $p$\% of all possible pairs.
The edges are established as follows:
First, we connect all successive nodes (\ie node 1\&2, 2\&3, ..., $n$\&1).
Then, we connect those separated by one node (\ie node 1\&3, 2\&4, ..., $n-1$\&1, $n$\&2), and afterwards, those separated by two nodes, three nodes, and so forth.
We continue this process until $p$\% are connected in total.
This leads to all nodes being connected to their local neighbors in a sliding window fashion.
Next, we turn $q$\% of the edges into outliers, \ie, random relative rotations.
We exclude the edges between successive nodes, so that every node gets at least two inlier edges.
Finally, all edges are perturbed by $\mathcal{N}(0, \sigma^2)$ and their order is randomized.
These edges are used as input to the rotation averaging algorithms.
The simulation is configured by setting $\{n, \ p, \ q, \ \sigma\}$ to one of the following values:
$n = \{100, 200\}$ rotations, $p = \{50, 20\}$\%, $q = \{0, 5, 10, ..., 50\}$\%, $\sigma = \{5, 10\}$ deg.
For each setting, we generate 100 independent datasets.

\begin{figure}[t]
    \centering
    \includegraphics[width=0.47\textwidth]{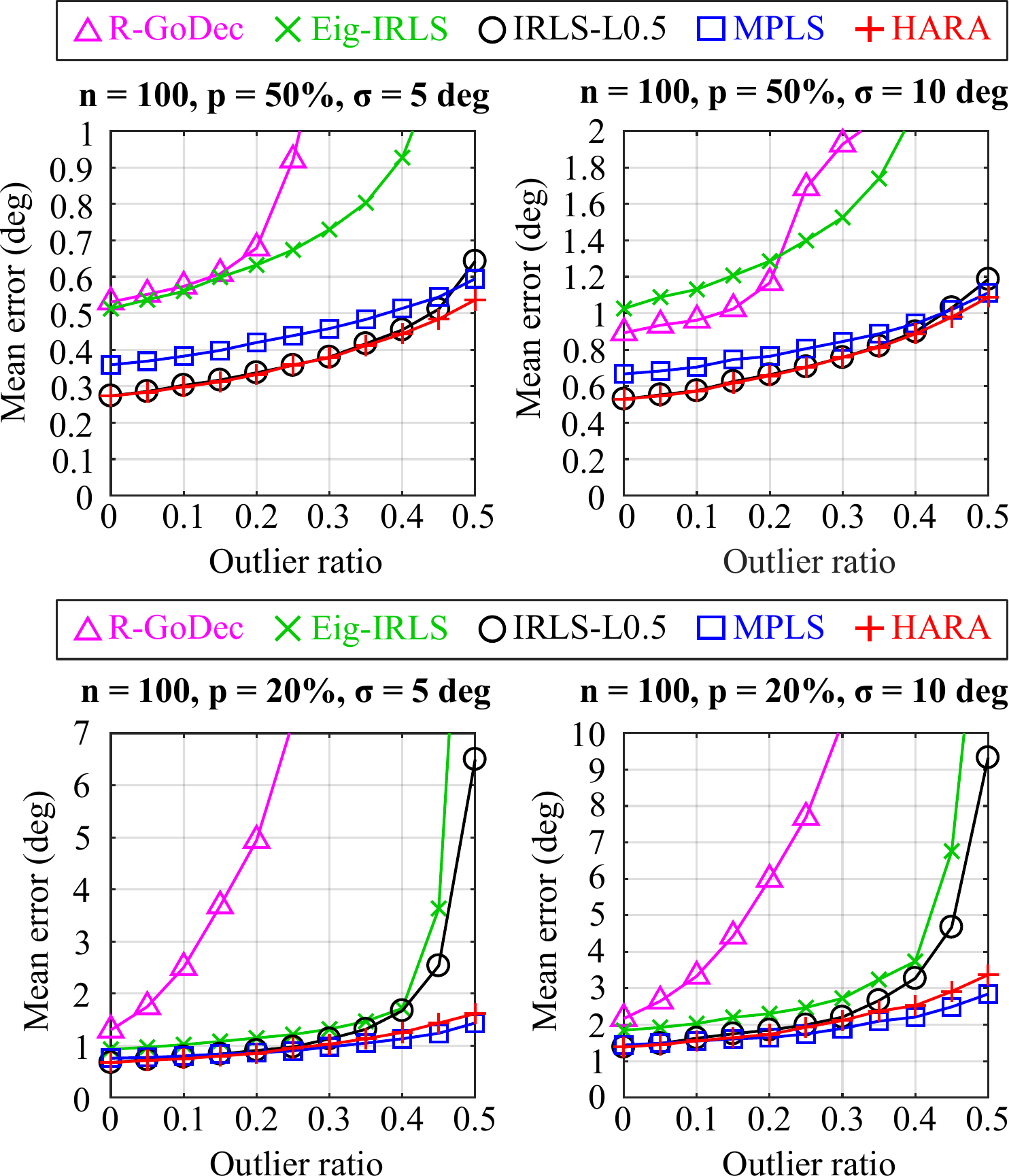}
    \caption{
    Simulation results (100 rotations):
    We plot the optimal mean errors, $\theta_1$ in Eq. \eqref{eq:theta_1}.
    For dense graphs ($p = 50$\%), IRLS-$\ell_{\frac{1}{2}}$, MPLS and HARA perform similarly well.
    For sparse graphs ($p = 20$\%), MPLS and HARA are more robust to outliers than the rest, with MPLS being slightly better at high outlier ratios.
    }
    \label{fig:synthetic_100}
    \vspace{-1em}
\end{figure}

\begin{figure}[t]
    \centering
    \includegraphics[width=0.47\textwidth]{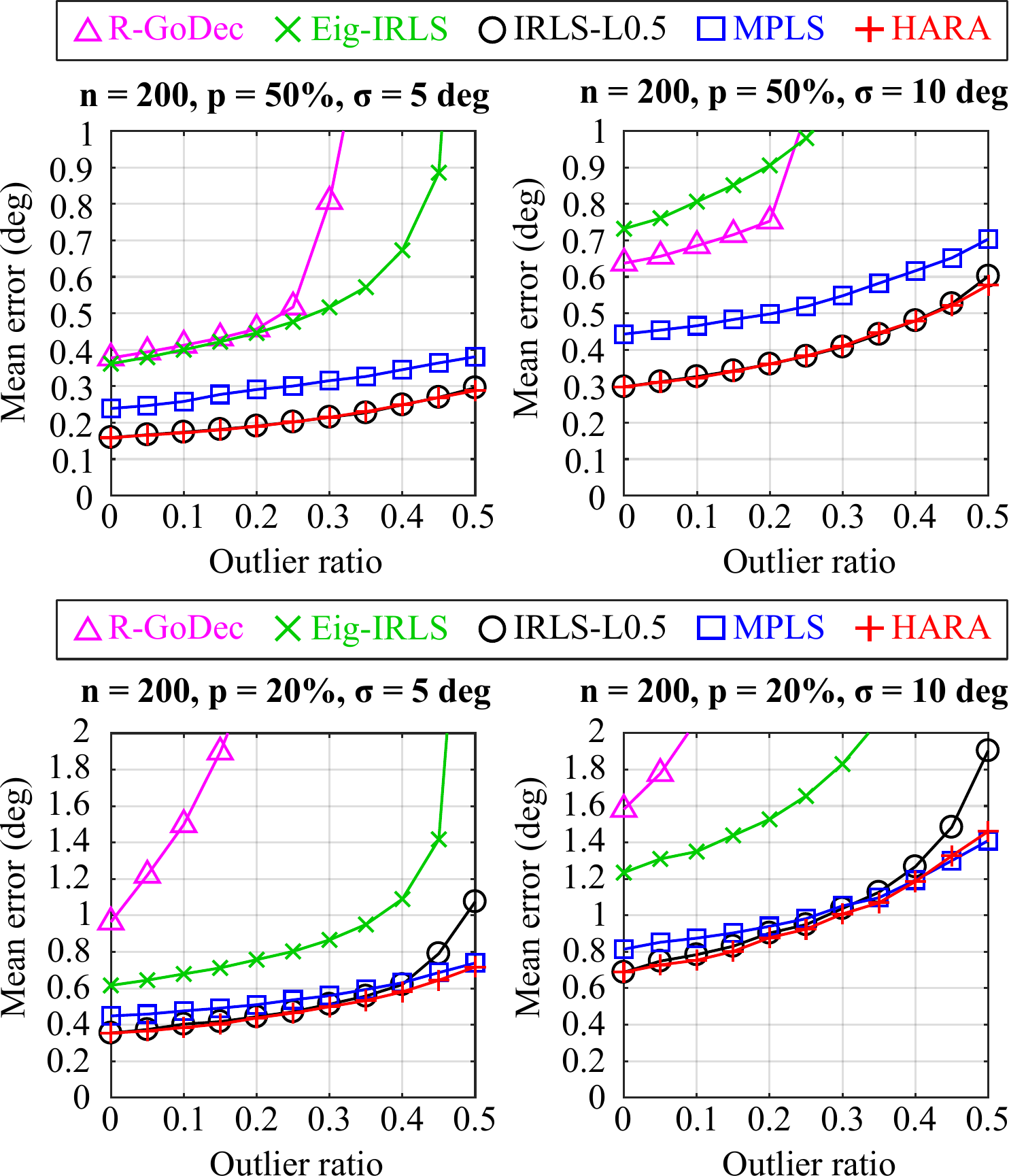}
    \caption{
    Simulation results (200 rotations):
    We plot the optimal mean errors, $\theta_1$ in Eq. \eqref{eq:theta_1}.
    IRLS-$\ell_{\frac{1}{2}}$, MPLS and HARA perform similarly well, except that for sparse graphs ($p = 20$\%), IRLS-$\ell_{\frac{1}{2}}$ is outperformed by the other two at high outlier ratios.
    }
    \label{fig:synthetic_200}
\end{figure}

Fig. \ref{fig:synthetic_100} and \ref{fig:synthetic_200} present the results for 100 and 200 rotations, respectively.
We see that MPLS and HARA are the two best performing methods, especially at high outlier ratios.


\newpage
\subsection{Real data}
\noindent\textbf{Without using the number of inlier feature matches:}\\
We evaluate the performance on the following real-world datasets: 1DSfM datasets\footnote{\url{http://www.cs.cornell.edu/projects/1dsfm/}\label{url:1dsfm_dataset}} \cite{wilson_2014_eccv}, `Notre Dame 715' (ND2) dataset\textsuperscript{\ref{url:chatterjee}} \cite{chatterjee_2018_tpami}, `Acropolis' (ACP), `Arts Quad' (ARQ) and `San Francisco' (SNF) datasets\footnote{\url{http://vision.soic.indiana.edu/projects/disco/}} \cite{crandall_2011_cvpr}.
As in \cite{chatterjee_2018_tpami}, only those cameras whose ground truth is available are used to evaluate the accuracy, even though those without the ground truth are still included in the input for rotation averaging.
Table \ref{tab:real_result1} reports the results.
It shows that in most cases HARA achieves state-of-the-art accuracy at a comparable speed.

\vspace{0.5em}
\noindent\textbf{Using the number of inlier feature matches:}\\
For this experiment, we only use the 1DSfM dataset\textsuperscript{\ref{url:1dsfm_dataset}} \cite{wilson_2014_eccv}, as the other datasets do not provide the 2D-2D correspondences.
The feature matches are only available for those cameras with the ground truth, so we disregard the rest.
To check the validity of the correspondences, we put a threshold (0.01) on the sine of the $L_1$-optimal angular reprojection error \cite{lee_2019_iccv, nee}. 
Table \ref{tab:real_result2} presents the results.
It shows that HARA achieves state-of-the-art results, with or without incorporating the number of inlier matches.

\section{Limitations}
\label{sec:limitations}
The main limitation of our method is that it is sensitive to the parameters we set, especially the loop thresholds.
Currently, we determine their values using a simple heuristic based on the sampled loop errors (Section \ref{subsec:initialization_full}).
We noticed that, in some of the real datasets, a small change in this heuristic introduces a non-negligible fluctuation in the initialization accuracy.
In future work, we plan to replace this heuristic with a more robust and reliable method. 

\section{Conclusion}
\label{sec:conclusion}
We presented HARA, a hierarchical approach for robust multiple rotation averaging.
For robust initialization of the rotation graph, we incrementally build a spanning tree based on a hierarchy of triplet support.
That is, the edges supported by many strong triplets are added in the tree sooner than those with fewer or weaker triplets. 
This approach significantly reduces the influence of outliers on the initial solution, allowing us to filter outliers prior to nonlinear optimization.
Also, we showed that we can optionally integrate the knowledge of the number of valid 2D-2D correspondences into our approach.
An extensive evaluation demonstrates that HARA achieves state-of-the-art results.

\clearpage
\newpage
\begin{table*}[t]
\begin{center}
\resizebox{\textwidth}{!}{%
\setlength{\tabcolsep}{3pt}
    \begin{tabular}{lrr|rrr|rrr|rrr|rrr|rrr|rrr}
    \thickhline
     \multicolumn{3}{c|}{\multirow{2}{*} {Datasets}}  & \multicolumn{3}{c|}{\multirow{2}{*} {R-GoDec \cite{arrigoni_2018_cviu}}}  & \multicolumn{3}{c|}{\multirow{2}{*}{Eig-IRLS \cite{arrigoni_2018_cviu}}}  & \multicolumn{3}{c|}{\multirow{2}{*} {IRLS-$\ell_{\frac{1}{2}}$ \cite{chatterjee_2018_tpami}}} & \multicolumn{3}{c|}{\multirow{2}{*}{MPLS$^{\bm{*}}$ \cite{shi_2020_icml}}} &\multicolumn{3}{c|}{\rule{0pt}{1em}Hybrid RA \cite{chen_2021_cvpr}}  & \multicolumn{3}{c}{{HARA w/o}}  \\
     & && && & && &&&& && & \multicolumn{3}{c|}{w/o VGF$\hspace{1pt}^{\bm{\dagger}}$} & \multicolumn{3}{c}{\#inlier matches} \\
    \thickhline
    \rule{0pt}{1em} \hspace{-0.5em} Name & \#views & \%edges& $\theta_1$ & $\theta_2$ & Time & $\theta_1$ & $\theta_2$ & Time & $\theta_1$ & $\theta_2$ & Time & $\theta_1$ & $\theta_2$ & Time & $\theta_1$ & $\theta_2$ & Time & $\theta_1$ & $\theta_2$ & Time \\

    ALM & 627 & 49.5\% & 6.3 & 16.3 & 4s & 3.9 & 12.2 & 21s & 4.2 & 12.6 & 27s & 3.7 & 12.1 & 29s & 4.3 & 12.7 & \multicolumn{1}{c|}{--} & {\color{blue}\textbf{3.5}} & {\color{blue}\textbf{11.5}} & 45s\\
    
    ELS & 247 & 66.8\%& 4.1 & 10.7 & 1s & 3.3 & 11.5 & 3s & 2.9 & 10.3 & 4s & 2.8 & 10.9 & 7s & 3.1 & 10.5 & \multicolumn{1}{c|}{--} &{\color{blue}\textbf{2.1}} & {\color{blue}\textbf{7.4}} & 7s\\
    
    GDM & 742 & 17.5\% & 51.3 & 64.8 & 17s & 65.8 & 74.2 & 29s & {\color{blue}\textbf{37.5}} & 62.3 & 12s & 40.7 & 68.7 & 74s & 44.7 & {\color{blue}\textbf{60.0}} & \multicolumn{1}{c|}{--} & 43.8 & 72.5 & 26s\\
    
    MDR & 394 & 30.7\% & 10.2 & 18.5 & 1s & 10.9 & 22.4 & 5s & 7.0 & 17.1 & 4s & 5.2 & 14.7 & 5s & 6.4 & 16.2 & \multicolumn{1}{c|}{--} & {\color{blue}\textbf{4.8}} & {\color{blue}\textbf{14.5}} & 14s\\
    
    MND & 474 & 46.8\% & 6.2 & 18.4 & 3s & 1.9 & 11.2 & 7s & 1.5 & 7.4 & 10s & 1.2 & 3.9 & 9s & 1.5 & 6.9 & \multicolumn{1}{c|}{--} & {\color{blue}\textbf{1.1}} & {\color{blue}\textbf{2.1}} & 20s\\
    
    ND1 & 553 & 68.1\% & 5.2 & 15.5 & 6s & 3.6 & 14.8 & 16s & 3.5 & 14.6 & 28s & 2.7 & 13.5 & 27s & 3.5 & 14.7 & \multicolumn{1}{c|}{--} & {\color{blue}\textbf{1.6}} & {\color{blue}\textbf{6.3}} & 55s\\
    
    NYC & 376 & 29.3\% & 6.4 & 9.9 & 6s & 3.8 & 8.2 & 5s & 3.0 & {\color{blue}\textbf{7.0}} &3s & 3.0 & 8.2 & 7s & 3.2 & 7.4 & \multicolumn{1}{c|}{--} & {\color{blue}\textbf{2.9}} & 7.7 & 10s\\
    
    PDP & 354 & 39.5\% & 11.4 & 22.0 & 2s & 4.0 & 9.3 & 8s & 4.1 & 8.1 & 4s & 3.5 & 8.2 & 4s & 5.3 & 10.4 & \multicolumn{1}{c|}{--} & {\color{blue}\textbf{3.4}} & {\color{blue}\textbf{7.4}} & 9s\\
    
    PIC & 2508 & 10.2\% & 24.8 & 40.0 & 150s & 81.0 & 91.2 & 687s & 6.8 & 18.6 & 467s & 4.6 & 14.6 & 295s & 7.0 & 20.1 & \multicolumn{1}{c|}{--} & {\color{blue}\textbf{4.4}} & {\color{blue}\textbf{13.1}} & 289s\\
    
    ROF & 1134 & 10.9\% & 12.6 & 19.5 & 61s & 3.4 & 10.4 & 52s & 3.1 & 10.2 & 12s & 2.8 & 10.0 & 13s & 3.1 & 9.2 & \multicolumn{1}{c|}{--} & {\color{blue}\textbf{2.7}} & {\color{blue}\textbf{8.5}} & 31s\\
    
    TOL & 508 & 18.5\% & 6.4 & 13.1 & 8s & 4.5 & 10.7 & 10s & {\color{blue}\textbf{3.9}} & {\color{blue}\textbf{9.0}} & 2s & 4.0 & 9.4 & 6s & 4.4 & 10.5 &\multicolumn{1}{c|}{--} & 4.3 & 10.0 & 13s\\
    
    TFG & 5433 & 4.6\% & 42.1 & 54.2 & 722s & 59.4 & 67.1 & 833s & 3.6 & {\color{blue}\textbf{9.8}} & 976s & 4.5 & 10.8 & 1945s & 15.1 & 17.8 & \multicolumn{1}{c|}{--} & {\color{blue}\textbf{3.5}} & 10.7 & 925s\\
    
    USQ & 930 & 5.9\% & 12.0 & 23.7 & 27s & 6.7 & 12.8 & 22s & 9.3 & 22.2 & 10s & 6.3 & 14.7 & 11s & 9.3 & 21.7 &\multicolumn{1}{c|}{--} & {\color{blue}\textbf{6.0}} & {\color{blue}\textbf{12.3}} & 9s \\
    
    VNC & 918 & 24.6\% & 16.1 & 36.9 & 25s & 8.6 & 28.4 & 27s & 8.3 & 27.5 & 28s & 6.2 & 18.2 & 53s & 8.4 & 27.2 & \multicolumn{1}{c|}{--} & {\color{blue}\textbf{6.1}} & {\color{blue}\textbf{18.1}} & 47s\\
    
    YKM & 458 & 26.5\% & 6.1 & 11.3 & 7s & 3.8 & 9.4 & 8s & 3.5 & 8.4 & 3s & 3.5 & 9.2 & 7s & 3.5 & 8.4 & \multicolumn{1}{c|}{--} & {\color{blue}\textbf{3.0}} & {\color{blue}\textbf{6.9}} & 16s\\
    
    ND2 & 715 & 25.3\% & 2.7 & 10.0 & 10s & 1.2 & 4.0 & 13s & {\color{blue}\textbf{1.1}} & {\color{blue}\textbf{3.5}} & 13s & {\color{blue}\textbf{1.1}} & 4.0 & 12s & {\color{blue}\textbf{1.1}} & {\color{blue}\textbf{3.5}} & \multicolumn{1}{c|}{--} & 1.3 & 5.5 & 23s\\
    
    ACP & 463 & 10.7\% & {\color{blue}\textbf{0.8}} & {\color{blue}\textbf{1.2}} & 6s & 1.1 & 1.7 & 4s & 1.2 & 1.7 & 1s & 1.4 & 2.0 & 2s & 1.2 & 1.7 & \multicolumn{1}{c|}{--} & 1.2 & 1.7 & 7s\\
    
    ARQ & 5530 & 1.5\% & 29.7 & 56.5 & 1111s & 70.1 & 79.7 & 4894s & 4.0 & 7.1 & 173s & {\color{blue}\textbf{3.2}} & {\color{blue}\textbf{6.3}} & 118s & 3.9 & 6.9 & \multicolumn{1}{c|}{--} & 3.6 & 6.8 & 137s\\
    
    SNF & 7866 & 0.3\% & \multicolumn{3}{c|}{Out of memory} & 77.3 & 87.3 & 3.8h & {\color{blue}\textbf{3.6}} & {\color{blue}\textbf{4.2}} & 180s & 4.4 & 5.5 & 154s & 4.3 & 6.2 & \multicolumn{1}{c|}{--} & {\color{blue}\textbf{3.6}} & {\color{blue}\textbf{4.2}} & 44s\\
    \thickhline
    \multicolumn{21}{c}{\rule{0pt}{1em} $\theta_1$ (deg): Optimal mean error in Eq. \eqref{eq:theta_1}, $\theta_2$ (deg): Optimal RMS error in Eq. \eqref{eq:theta_2}, $\text{\%edges} = \text{\#edges}/\text{\#possible pairs of views}$ in \%. }\\
    \multicolumn{21}{c}{\rule{0pt}{1em} $^{\bm{*}}$Due to the non-deterministic nature of MPLS \cite{shi_2020_icml}, we report the median of five independent runs. }\\
    \multicolumn{21}{c}{\rule{0pt}{1em} $^{\bm{\dagger}}$The computation times of Hybrid RA \cite{chen_2021_cvpr} are not included for comparison, since it is the only method implemented in C++. }\\
    \thickhline
    \end{tabular}%
    }
    \vspace{-1.5em}
    \end{center}
    \caption{
    Results on the real datasets \textbf{without} the knowledge of the 2D-2D correspondences: 
    For all datasets, HARA gives either better or comparable results to the state of the art.
    Interestingly, for the SNF dataset, it takes substantially less time than the rest.
    The results of MPLS \cite{shi_2020_icml} are mostly competitive with ours, except for the ELS, ND1, TFG and SNF datasets where HARA performs noticeably better.
    We run Hybrid RA \cite{chen_2021_cvpr} without view graph filtering because this requires the number of valid 2D-2D correspondences.
    As a result, this method does not provide much gain in accuracy compared to IRLS-$\ell_{\frac{1}{2}}$ \cite{chatterjee_2018_tpami}, even though it performs an additional global optimization prior to local refinement.
    In fact, Hybrid RA performs much worse than IRLS-$\ell_{\frac{1}{2}}$ on the TFG dataset. 
    } 
    \label{tab:real_result1}
\end{table*}

\begin{table*}[t]
\begin{center}
\resizebox{\textwidth}{!}{%
\setlength{\tabcolsep}{3pt}
    \begin{tabular}{lrr|rrr|rrr|rrr|rrr|rrr|rrr}
    \thickhline
     \multicolumn{3}{c|}{\multirow{2}{*} {Datasets}}  & \multicolumn{3}{c|}{\multirow{2}{*} {IRLS-$\ell_{\frac{1}{2}}$ \cite{chatterjee_2018_tpami}}}  & \multicolumn{3}{c|}{\multirow{2}{*}{MPLS$^{\bm{*}}$ \cite{shi_2020_icml}}}  & \multicolumn{3}{c|}{\rule{0pt}{1em}Hybrid RA \cite{chen_2021_cvpr}} & \multicolumn{3}{c|}{\rule{0pt}{1em}Hybrid RA \cite{chen_2021_cvpr}} &\multicolumn{3}{c|}{\rule{0pt}{1em} HARA w/o}  & \multicolumn{3}{c}{HARA with}  \\
     & && &&&&& &\multicolumn{3}{c|}{w/o VGF$\hspace{1pt}^{\bm{\dagger}}$} &\multicolumn{3}{c|}{with VGF$\hspace{1pt}^{\bm{\dagger}}$} &\multicolumn{3}{c|}{\#inlier matches} &\multicolumn{3}{c}{\#inlier matches} \\
    \thickhline
    \rule{0pt}{1em} \hspace{-0.5em} Name & \#views & \%edges& $\theta_1$ & $\theta_2$ & Time & $\theta_1$ & $\theta_2$ & Time & $\theta_1$ & $\theta_2$ & Time & $\theta_1$ & $\theta_2$ & Time & $\theta_1$ & $\theta_2$ & Time & $\theta_1$ & $\theta_2$ & Time \\

    ALM & 577 & 58.4\% & 4.0 & 12.4 & 28s & 3.7 & 12.0 & 24s & 4.3 & 12.8 & \multicolumn{1}{c|}{--} & {\color{blue}\textbf{3.0}} & {\color{blue}\textbf{10.2}} & \multicolumn{1}{c|}{--} & 3.5 & 11.5 & 41s & 3.4 & 11.0 & 40s\\

    ELS & 227 & 78.0\% & 2.8 & 10.1 & 2s & 3.0 & 11.7 & 6s & 3.0 & 9.9 & \multicolumn{1}{c|}{--} & 2.1 & 7.1 & \multicolumn{1}{c|}{--} & 2.1 & 7.2 & 8s & {\color{blue}\textbf{1.8}} & {\color{blue}\textbf{4.8}} & 6s\\
    
    GDM & 677 & 20.9\% & 37.4 & 62.2 & 8s & 40.7 & 68.6 & 81s & {\color{blue}\textbf{34.5}} & {\color{blue}\textbf{55.3}} & \multicolumn{1}{c|}{--} & 39.9 & 68.7 & \multicolumn{1}{c|}{--} & 44.1 & 72.5 & 23s & 44.3 & 72.8 & 16s\\

    MDR & 341 & 40.7\% & 6.7 & 16.7 & 4s & 5.1 & 14.4 & 4s & 6.3 & 15.8 & \multicolumn{1}{c|}{--} & {\color{blue}\textbf{4.4}} & {\color{blue}\textbf{13.1}} & \multicolumn{1}{c|}{--} & 4.8 & 14.5 & 13s & 4.8 & 14.8 & 11s\\
    
    MND & 450 & 51.8\% & 1.5 & 7.4 & 6s & 1.2 & 3.9 & 8s & 1.5 & 6.9 & \multicolumn{1}{c|}{--} & {\color{blue}\textbf{1.1}} & 2.2 & \multicolumn{1}{c|}{--} & {\color{blue}\textbf{1.1}} & {\color{blue}\textbf{2.1}} & 21s & {\color{blue}\textbf{1.1}} & {\color{blue}\textbf{2.1}}  & 18s\\
    
    ND1 & 553 & 68.1\% & 3.5 & 14.6 & 29s & 2.8 & 13.6 & 30s & 3.5 & 14.7 & \multicolumn{1}{c|}{--} & 1.7 & 6.1 & \multicolumn{1}{c|}{--} & 1.6 & 6.3 & 61s & {\color{blue}\textbf{1.5}} & {\color{blue}\textbf{5.9}} & 44s\\
    
    NYC & 332 & 37.4\% & 3.1 & 7.1 & 3s & 3.1 & 8.2 & 7s & 3.4 & 7.8 & \multicolumn{1}{c|}{--} & 3.0 & 7.1 &\multicolumn{1}{c|}{--}  & 2.9 & 7.8 & 8s & {\color{blue}\textbf{2.6}} & {\color{blue}\textbf{5.8}} & 7s\\
    
    PDP & 338 & 43.3\% & 4.1 & 8.2 & 4s & 3.5 & 8.2 & 4s & 5.2 & 10.3 & \multicolumn{1}{c|}{--} & {\color{blue}\textbf{3.1}} & {\color{blue}\textbf{6.4}} & \multicolumn{1}{c|}{--} & 3.5 & 7.8 & 8s & 3.3 & 6.6 & 7s\\
    
    PIC & 2152 & 13.4\% & 6.2 & 17.0 & 419 & 4.7 & 14.6 & 254s & 6.3 & 18.6 & \multicolumn{1}{c|}{--} & 4.3 & 12.1 & \multicolumn{1}{c|}{--} & 4.1& {\color{blue}\textbf{11.3}} & 269s & {\color{blue}\textbf{4.0}} & {\color{blue}\textbf{11.3}} & 247s\\
    
    ROF & 1084 & 11.9\% & 3.1 & 10.2 & 16s & 2.8 & 9.7 & 13s & 3.1 & 9.4 & \multicolumn{1}{c|}{--} & {\color{blue}\textbf{2.5}} & {\color{blue}\textbf{6.7}} & \multicolumn{1}{c|}{--} & 2.7 & 8.7 & 30s & {\color{blue}\textbf{2.5}} & 7.6 & 25s\\
    
    TOL & 472 & 21.4\% & {\color{blue}\textbf{3.9}} & {\color{blue}\textbf{8.9}} & 2s & 4.0 & 9.4 & 4s & 4.4 & 10.4 & \multicolumn{1}{c|}{--} & 4.0 & 9.4 & \multicolumn{1}{c|}{--} & 4.3 & 10.0 & 8s & 4.0 & {\color{blue}\textbf{8.9}} & 11s\\
    
    TFG & 5058 & 5.3\% & 3.5 & {\color{blue}\textbf{8.9}} & 881s & 5.3 & 11.1 & 1466s & 4.0 & 9.8 & \multicolumn{1}{c|}{--} & 5.3 & 11.8 & \multicolumn{1}{c|}{--} & 3.5 & 10.1 & 948s & {\color{blue}\textbf{3.4}} & 9.6 & 902s\\
    
    USQ & 789 & 7.9\% & 6.7 & 14.2 & 5s & 6.2 & 12.9 & 7s & 7.9 & 17.0 & \multicolumn{1}{c|}{--} & 6.6 & 14.8 & \multicolumn{1}{c|}{--}  & 5.9 & 11.2 & 10s & {\color{blue}\textbf{5.8}} & {\color{blue}\textbf{10.8}} & 9s\\
    
    VNC & 836 & 29.6\% & 8.4 & 27.5 & 32s & {\color{blue}\textbf{6.2}} & 18.2 & 59s & 8.4 & 27.1 & \multicolumn{1}{c|}{--} & 6.3 & {\color{blue}\textbf{18.0}} & \multicolumn{1}{c|}{--} & {\color{blue}\textbf{6.2}} & 18.1 & 54s & {\color{blue}\textbf{6.2}} & 18.1 & 52s\\
    
    YKM & 437 & 29.1\% & 3.5 & 8.4 & 2s & 3.6 & 9.4 & 4s & 3.6 & 8.4 & \multicolumn{1}{c|}{--} & 3.9 & 11.8 & \multicolumn{1}{c|}{--} & {\color{blue}\textbf{3.0}} & {\color{blue}\textbf{6.9}} & 11s & 3.4 & 11.2 & 12s\\
    
    \thickhline
    \multicolumn{21}{c}{\rule{0pt}{1em} $\theta_1$ (deg): Optimal mean error in Eq. \eqref{eq:theta_1}, $\theta_2$ (deg): Optimal RMS error in Eq. \eqref{eq:theta_2}, $\text{\%edges} = \text{\#edges}/\text{\#possible pairs of views}$ in \%. }\\
    \multicolumn{21}{c}{\rule{0pt}{1em} $^{\bm{*}}$Due to the non-deterministic nature of MPLS \cite{shi_2020_icml}, we report the median of five independent runs. }\\
    \multicolumn{21}{c}{\rule{0pt}{1em} $^{\bm{\dagger}}$The computation times of Hybrid RA \cite{chen_2021_cvpr} are not included for comparison, since it is the only method implemented in C++. }\\
    \thickhline
    \end{tabular}%
    }
    \vspace{-1.5em}
    \end{center}
    \caption{
    Results on the real datasets \textbf{with} the knowledge of the number of valid 2D-2D correspondences: 
    The best performing methods are Hybrid RA \cite{chen_2021_cvpr} (with VGF) and HARA with and without using \#inlier feature matches.
    All three of these methods give competitive results, but on the TFG and USQ datasets, HARA outperforms Hybrid RA by a noticeable margin.
    } 
    \label{tab:real_result2}
\end{table*}

\clearpage
\newpage
{\small
\balance
\bibliographystyle{ieee_fullname}

}

\end{document}